\pdfoutput=1
\documentclass[11pt]{article}

\usepackage[final]{acl}

\usepackage{times}
\usepackage{latexsym}

\usepackage[T1]{fontenc}
\usepackage[utf8]{inputenc}

\usepackage{microtype}

\usepackage{inconsolata}

\usepackage{lipsum}
\usepackage{comment}
\usepackage{tabularx}
\usepackage{color, colortbl}
\usepackage{fontawesome5}
\usepackage{fdsymbol}
\usepackage[cachedir=., draft=true]{minted}
\usepackage{graphicx}

\title{Indexing Portuguese NLP Resources with PT-Pump-Up}

\author{
  Rúben Almeida \\
  INESC TEC \\
  \texttt{\bf ruben.f.almeida@inesctec.pt} \\ \And
  Ricardo Campos \\
  INESC TEC, Uni. Beira Interior \\
  \texttt{\bf ricardo.campos@ubi.pt} \\ \AND
  Alípio Jorge \\
  INESC TEC, Uni. of Porto \\
  \texttt{\bf amjorge@fc.up.pt} \\ \And
  Sérgio Nunes \\
  INESC TEC, Uni. of Porto\\
  \texttt{\bf ssn@fe.up.pt}
}

\begin{document}
\maketitle

\begin{abstract}

The recent advances in natural language processing (NLP) are linked to training processes that require vast amounts of corpora. Access to this data is commonly not a trivial process due to resource dispersion and the need to maintain these infrastructures online and up-to-date. New developments in NLP are often compromised due to the scarcity of data or lack of a shared repository that works as an entry point to the community. This is especially true in low and mid-resource languages, such as Portuguese, which lack data and proper resource management infrastructures. In this work, we propose PT-Pump-Up, a set of tools that aim to reduce resource dispersion and improve the accessibility to Portuguese NLP resources. Our proposal is divided into four software components: a)~\href{http://pt-pump-up.inesctec.pt/}{a web platform} to list the available resources; b)~\href{https://pypi.org/project/pt-pump-up/}{a client-side Python package} to simplify the loading of Portuguese NLP resources; c)~\href{https://pypi.org/project/pt-pump-up-admin/}{an administrative Python package} to manage the platform and d)~\href{https://github.com/LIAAD/PT-Pump-Up/wiki}{a public GitHub repository} to foster future collaboration and contributions. All four components are accessible using:~\url{https://linktr.ee/pt_pump_up}

\end{abstract}

\section{Introduction}

The topic of NLP resource management in European languages was initially introduced by~\citet{aei234726}, with the first references to Portuguese resources presented ten years later in the works of~\citet{santos2002centro}. The recent advances in NLP, linked to the development of large language models reintroduced the debate about NLP resource management due to the large volume of training data required by these architectures. Several platforms have been recenlty introduced offering different approaches to address this problem. In Table~\ref{tab:platforms}, we present some of these platforms focusing on those that index, to some extent, Portuguese NLP resources. Our analysis reveals that Portuguese NLP resources are dispersed across more than 11 platforms which implement different storage policies; some only store metadata, while others focus on providing entire copies of the resources indexed. In a mid-resource language such as Portuguese~\cite{DBLP:journals/corr/abs-2004-09095}, this resource dispersion phenomenon exacerbates the already existing challenges linked to the reduced amount of NLP resources, negatively impacting the accessibility to these resources.

To address these challenges, we extend the surveying works of~\citet{de2023building} and propose PT-Pump-Up, a set of tools that support the development of the first centralising platform for Portuguese NLP resources. In this demonstration, we present the minimum set of valuable features to achieve this goal divided across the four software components that compose PT-Pump-Up: a)~\href{http://pt-pump-up.inesctec.pt/}{a web platform}; b)~\href{https://pypi.org/project/pt-pump-up/}{a client Python package}; c)~\href{https://pypi.org/project/pt-pump-up-admin/}{an administrative Python package} and d)~\href{https://github.com/LIAAD/PT-Pump-Up}{a public GitHub repository}. Additional details about this release are available in the \href{https://github.com/LIAAD/PT-Pump-Up/wiki}{wiki of the project}.

\begin{table*}[ht]
\centering
\begin{tabularx}{\linewidth}{Xccccccc}
\hline
\textbf{Platform} & \textbf{Updated} & \textbf{Origin} & \textbf{\#~PT Var.} & \textbf{Colab.} & \textbf{Meta.} & \textbf{Res.} \\ \hline

\href{https://portulanclarin.net/}{Portulan Clarin}~\cite{branco2023clarin}  & $\checkmark$ & PT & 1 & \faExclamationTriangle & \faExclamationTriangle & $\checkmark$ \\ \hline

\href{http://www.nilc.icmc.usp.br/nilc/index.php/tools-and-resources}{NILC: Tools and Resources} & $\checkmark$ & BR & 1 & $\times$ & $\times$ & $\checkmark$ \\ \hline

\href{https://github.com/ajdavidl/Portuguese-NLP}{Portuguese-NLP} & $\checkmark$ & BR & \faExclamationTriangle & $\checkmark$ & $\checkmark$ & $\times$ \\ \hline

\href{https://huggingface.co/}{HuggingFace} & $\checkmark$ & FR & \faExclamationTriangle & $\checkmark$ & \faExclamationTriangle & $\checkmark$\\ \hline

\href{https://paperswithcode.com/}{PapersWithCode} & $\checkmark$ & USA & \faExclamationTriangle & $\checkmark$ & $\checkmark$ & $\times$ \\ \hline

\href{https://catalogue.elra.info/en-us/}{ELRA}& $\checkmark$ & BE & 1 & $\times$ & $\times$ & $\checkmark$ \\ \hline

\href{http://www.language-archives.org/index.html}{Open Language Archives Community}~\cite{simons2003open} & $\checkmark$ & USA & 2+ & $\times$ & $\checkmark$ & $\times$ \\ \hline

\rowcolor{lightgray!30} % To be changed in examples

\href{https://forum.ailab.unb.br/t/datasets-em-portugues/251}{AiLab} & 2021 & BR & \faExclamationTriangle & $\checkmark$ & $\checkmark$ & $\times$ \\ \hline

\rowcolor{lightgray!30} % To be changed in examples

\href{https://aclweb.org/aclwiki/Resources_for_Portugese}{ACL Wiki: Resources for Portuguese} & 2020 & USA & \faExclamationTriangle & $\times$ & $\checkmark$ & $\times$ \\ \hline

\rowcolor{lightgray!30} % To be changed in examples

\href{https://linguistic-datasets-pt.etica.ai/}{Organização Etica.AI} &  2018 & BR & \faExclamationTriangle & $\times$ & $\checkmark$ & $\times$ \\\hline

\rowcolor{lightgray!30} % To be changed in examples

\href{https://www.linguateca.pt/}{Linguateca}~\cite{santos2004linguateca} & 2012 & PT & \faExclamationTriangle & $\times$ & $\times$ & $\checkmark$ \\ \hline \hline

\rowcolor{cyan!05} % To be changed in examples
\href{http://pt-pump-up.inesctec.pt/}{PT-Pump-Up} & $\checkmark$ & PT & 2+ & $\checkmark$ & $\checkmark$ & \faExclamationTriangle \\ \hline

\hline
\end{tabularx}

\caption{Overview of NLP resource management platforms supporting Portuguese resources, emphasizing discontinued projects (shaded in grey). Details encompass platform origin, collaborative operation (Colab.), and storage of resource metadata (Meta.) or complete copies of resources (Res.). The warning sign indicates omitted or incomplete information. For the count of Portuguese Varieties indexed (\#PT Var), "2+" denotes platforms considering at least the two predominant Portuguese varieties: European and Brazilian.}

\label{tab:platforms}
\end{table*}

\section{PT-Pump-Up}

\begin{figure*}[ht]
    \centering
    \includegraphics[width=0.9\linewidth]{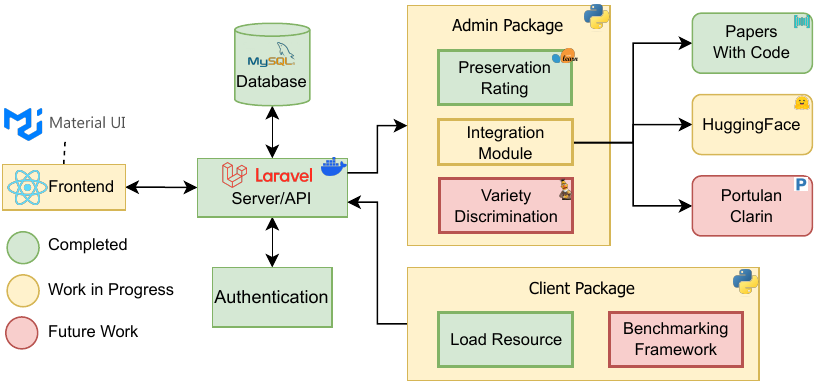}
    \caption{Architecture of PT-Pump-Up. Background colours highlight the completeness of each module.}
    \label{fig:pt_pump_up_arch}
\end{figure*}

The architecture of PT-Pump-Up is represented in Figure~\ref{fig:pt_pump_up_arch}, which highlights not only the features already implemented but also the work in progress and future plans associated with this project. In this demonstration, we present four scenarios where PT-Pump-Up can be employed to mitigate resource dispersion and enhance synchronization across diverse platforms that support Portuguese NLP resources.

\subsection{Easy Access to Portuguese NLP Resources}

The PT-Pump-Up Python client\footnote{\url{https://pypi.org/project/pt-pump-up}} permits the easy loading of Portuguese NLP resources. The resource is loaded directly if it has a copy in HuggingFace\footnote{\url{https://huggingface.co}}; if not, it returns the metadata that describes it. In Listing~\ref{code:load_dataset}, we demo how to use PT-Pump-Up to achieve this goal using a few lines of code.

\begin{listing}[!h]
    \begin{minted}[linenos=false, breaklines, breakafter=d, fontsize=\small, xleftmargin=-15pt]{python}
    from pt_pump_up.PT_Pump_Up import PTPumpUp
    client = PTPumpUp()
    all_ner_datasets = client.all_datasets(nlp_task="Named Entity Recognition")
    print(all_ner_datasets.head())
    # Dataset is Loaded as a HF Dataset object
    dataset = client.load_dataset(english_name=str)
    \end{minted}
    \vspace{-15pt}
    \caption{Load Portuguese named entity recognition dataset.}
    \label{code:load_dataset}
\end{listing}

\subsection{Indexing Portuguese NLP Resource}

The PT-Pump-Up \href{https://pypi.org/project/pt-pump-up-admin/}{administrative package} permits authenticated \href{https://pt.wikipedia.org/wiki/CRUD}{CRUD operations} to manage the resources indexed in the platform. These actions can also be done using the \href{http://pt-pump-up.inesctec.pt/}{web interface}, ensuring that the absence of programming skills is not a barrier to interacting with the platform. In Listing~\ref{code:crud_operations}, we demonstrate how a new NLP task can be included in PT-Pump-Up with a few lines of Python code.

\begin{listing}[!h]
    \begin{minted}[linenos=false, breaklines, breakafter=d, fontsize=\small, xleftmargin=-15pt]{python}
    from pt_pump_up_admin.PT_Pump_Up import PT_Pump_Up
    from pt_pump_up_admin.crud.NLPTask import NLPTask
    
    pt_pump_up = PT_Pump_Up(
        # Get token: pt-pump-up.inesctec.pt/dashboard
        bearer_token=str,
    )
    NLPTask(pt_pump_up=pt_pump_up).insert(
        name=str,
        acronym=str,
        papers_with_code_ids=list,
    )
    \end{minted}
    \vspace{-15pt}
    \caption{Inserting a NLP task to the database.}
    \label{code:crud_operations}
\end{listing}

\subsection{Measure Resource Preservation Needs}

We introduced a \textit{resource preservation rating} to identify less accessible resources. This metric enabled the establishment of a hybrid storage policy. For resources with high preservation ratings, only metadata is stored, while those with lower ratings are prioritized for human intervention and the creation of a backup copy. The preservation rating can be provided during resource submission or automatically determined using a decision tree integrated into the \href{https://pypi.org/project/pt-pump-up-admin/}{admin package}.

\begin{listing}[!h]
    \begin{minted}[linenos=false, breaklines, breakafter=d, fontsize=\small, xleftmargin=-15pt]{python}    
    from pt_pump_up.rating.Preservation_Rating import PreservationRatingDataset
    
    # Instantiates a client
    client = PreservationRatingDataset()
    
    preservation_rating = client.predict(#...dataset proprieties)
    print(preservation_rating)
    \end{minted}
    \caption{Predicting preservation rating of a dataset based on its metadata}
    \label{code:preservation_rating}
\end{listing}

\subsection{Integrate With Papers With Code}

The PT-Pump-Up integration module included in the admin package compresses the logic developed to enforce resource synchronization with other platforms. In this release, we deliver the tools to support the integration with \href{https://paperswithcode.com/}{Papers With Code}. This module presents challenges due to the heterogeneity of systems used by the targeted applications. In Listing~\ref{code:papers}, we demonstrate how PT-Pump-Up can be used to synchronise a resource with Papers With Code using a few lines of code.

\begin{listing}[!h]
    \begin{minted}[linenos=false, breaklines, breakafter=d, fontsize=\small, xleftmargin=-15pt]{python}
    from pt_pump_up.papers_with_code.PapersWithCode import PapersWithCodeDataset, PapersWithCode
    
    # Login in PapersWithCode
    client = PapersWithCode(username=str, password=str)
    
    #Create Dataset instance
    dataset = PapersWithCodeDataset(#...dataset proprieties)
    #Publish Resource
    client.insert(dataset)
    \end{minted}
    \caption{Insert dataset metadata to Papers With Code.}
    \label{code:papers}
\end{listing}

\section{Conclusion and Future Work}

This paper details the first release of PT-Pump-Up and how its tools can be used to address the challenge of Portuguese NLP resource dispersion. In this release, we deliver the minimum valuable set of essential features capable of demonstrating the four software modules that compose PT-Pump-Up. This project is a work in progress, with future work focusing on extending the integration module to other platforms and extending the resources indexed with the collaboration of the Portuguese NLP community.

\section*{Acknowledgements}
This work is financed by National Funds through the FCT - Fundação para a Ciência e a Tecnologia, I.P. within the project StorySense, reference 2022.09312.PTDC(DOI 10.54499/2022.09312.PTDC).

\bibliography{custom}

\end{document}